\documentclass[review,12pt,authoryear]{elsarticle}

\usepackage{epstopdf} 
\usepackage{fancyhdr, lastpage}
\usepackage{todonotes}
\usepackage[vlined,algoruled,commentsnumbered]{algorithm2e}
\usepackage{amsmath} 
\usepackage{listings}
\usepackage{subfigure}
\usepackage{hyperref}
\usepackage{soul}
\usepackage{algorithmic}
\usepackage{caption}

\usepackage{caption} 
\let\oldnl\nl
\newcommand{\nonl}{\renewcommand{\nl}{\let\nl\oldnl}}

\newlength\lenKwIn

\newlength\lenKwOut

\SetKwData{ms}{$M$}           
\SetKwData{pop}{$pop$}
\SetKwData{popI}{$population\_I$}
\SetKwData{popII}{$population\_II$}
\SetKwData{interpop}{$interPop$}
\SetKwData{gen}{$gen$}
\SetKwData{maxgens}{$G$}
\SetKwData{parent}{$parent$}
\SetKwData{parentI}{$parent_1$}
\SetKwData{parentII}{$parent_2$}
\SetKwData{child}{$child$}
\SetKwData{childI}{$child_1$}
\SetKwData{childII}{$child_2$}

\SetKwFunction{init}{init}
\SetKwFunction{eval}{evaluate}
\SetKwFunction{size}{size}
\SetKwFunction{selectT}{selection}
\SetKwFunction{merge}{merge}
\SetKwFunction{cross}{mate}
\SetKwFunction{mutate}{mutate}
\SetKwFunction{rand}{random}
\SetKwFunction{add}{add}
\SetKwFunction{getbest}{getBestSolutions}

\newcommand{\phaseI} {\texttt{phase\text{\,}I}}
\newcommand{\phaseII} {\texttt{phase\text{\,}II}}
\newcommand{\distI} {$dist_R$}
\newcommand{\dist} {$dist$}
\newcommand{\sr} {$sat$}

\newcommand{\prodgraph} {$G$}
\newcommand{\countries} {$C$}
\newcommand{\psites} {$L$}      
\newcommand{\pplants} {$F$}     
\newcommand{\punits} {$U$}      
\newcommand{\suppliers} {$S$}
\newcommand{\parts} {$P$}
\newcommand{\warehouses} {$W$}
\newcommand{\punitsc} {$U^c$}      
\newcommand{\punitsp} {$U^p$}      
\newcommand{\punitss} {$U^s$}      
\newcommand{\warehousesu} {$W^u$}  
\newcommand{\va} {$v^{p}$}
\newcommand{\vac} {$v^{c}$}      
\newcommand{\vas} {$v^{s}$}      
\newcommand{\vacmin} {$v^{c}_{min}$}
\newcommand{\vacmax} {$v^{c}_{max}$}
\newcommand{\vasmin} {$v^{s}_{min}$}
\newcommand{\vasmax} {$v^{s}_{max}$}

\newcommand{\vaumax} {$v^{u}_{max}$}
\newcommand{\trmeans} {$T$}       
\newcommand{\tremission} {$t_c$}  
\newcommand{\trspeed} {$t_s$}     
\newcommand{\trcost} {$t_t$}     
\newcommand{\trload} {$t_l$}      
\newcommand{\trmgraph} {$\mathcal{G}^{T}$}   
\newcommand{\indsys} {$I$}        

\usepackage{verbatim}

\journal{Expert Systems with Applications}

\begin{document}
\begin{frontmatter}



\title{Towards Resilient and Sustainable Global Industrial Systems: An Evolutionary-Based Approach} 

\author{V\'{a}clav Jirkovsk\'{y}\fnref{ciirc}\corref{cor1}}
\ead{vaclav.jirkovsky(at)cvut.cz}
\author{Ji\v{r}\'{i} Kubal\'{i}k\fnref{ciirc}}
\ead{jiri.kubalik(at)cvut.cz}
\author{Petr Kadera\fnref{ciirc}}
\ead{petr.kadera(at)cvut.cz}
\author{Arnd Schirrmann\fnref{airbus}}
\ead{Arnd.Schirrmann(at)airbus.com}
\author{Andreas Mitschke\fnref{airbus}}
\ead{andreas.a.mitschke(at)airbus.com}
\author{Andreas Zindel\fnref{airbus}}
\ead{andreas.zindel(at)airbus.com}

\cortext[cor1]{Corresponding author.}
\cortext[cor2]{This work was co-funded by the European Union under the project
Robotics and advanced industrial production (reg. no.
CZ.02.01.01/00/22\_008/0004590) and by institutional resources for research by the Czech Technical University in Prague, Czech Republic.}
\affiliation[ciirc]{organization={Czech Institute of
Informatics, Robotics and Cybernetics, Czech Technical University in
Prague},
            city={Prague},
            postcode={16000}, 
            country={Czech Republic}}

\affiliation[airbus]{organization={Airbus Central R\&T},
            country={Germany}}

\begin{abstract}
This paper presents a new complex optimization problem in the field of automatic design of advanced industrial systems and proposes a hybrid optimization approach to solve the problem. The problem is multi-objective as it aims at finding solutions that minimize CO$_2$ emissions, transportation time, and costs. The optimization approach combines an evolutionary algorithm and classical mathematical programming to design resilient and sustainable global manufacturing networks. Further, it makes use of the OWL ontology for data consistency and constraint management.
The experimental validation demonstrates the effectiveness of the approach in both single and double sourcing scenarios.
The proposed methodology, in general, can be applied to any industry case with complex manufacturing and supply chain challenges.
\end{abstract}

\begin{graphicalabstract}
\includegraphics[width=0.99\linewidth]{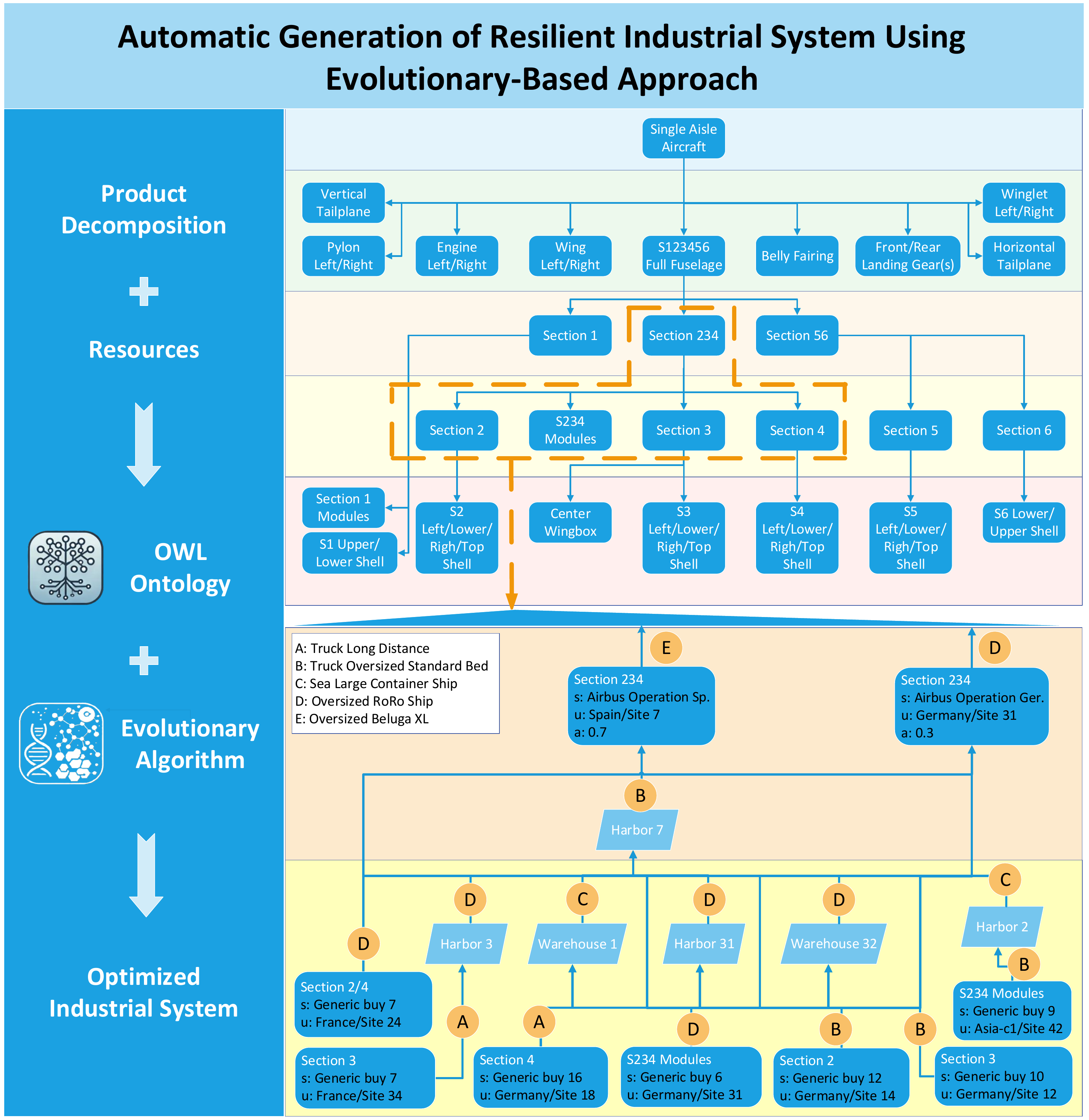}
\end{graphicalabstract}

\begin{highlights}
    \item The formalization of an optimization problem for global industrial system design.
    \item Optimization integrating an evolutionary algorithm and mathematical programming. 
    \item Use of OWL ontology for data consistency and constraint management.
\end{highlights}

\begin{keyword}
multi-objective optimization \sep evolutionary algorithm \sep ontology \sep industrial system \sep supply chain



\end{keyword}

\end{frontmatter}



\section{Introduction}
\label{sec:introduction}

This paper presents the results of research work performed at a large European aircraft
manufacturer that has focused on the capabilities development and verification for the design of novel, disruptive manufacturing systems for future aircraft programs.

The production of commercial aircraft is a complex and highly distributed process. For example, the Airbus A380 comprises 2.5 million individual parts, produced by 1500 companies in 30 countries worldwide~\citep{Airbus2021}. Due to the need for highly specialized parts and strict certification requirements (concerning safety and dual use) production facilities for
these parts cannot be created on a green field. Rather these parts are sourced from
established, advanced manufacturers both from company owned subsidiaries and external
suppliers. Thus both the optimization of production location allocations and transportation of the produced parts is pivotal to the performance of the industrial system for creating such complex products.

Efficient transportation of the larger parts of an aircraft is crucial. Major components (MCs) such as wings, engines, sections, etc. cannot be transported with regular shipping methods. The aforementioned A380 has a wing span of almost 79.8\,m with an overall length of 72.7\,m and a fuselage diameter of 7.1\,m~\citep{Airbus2021}. In comparison a standard 40\,ft container with dimensions of 12.19\,m length, 2.44\,m width, 2.59\,m height is only suitable for smaller parts of the aircraft. Given that some parts can only be manufactured at specific locations puts major constraints onto the design of a global industrial system architecture.

To transport these major components, special transportation means are required (oversized
road transporters with or without obligatory escorts, large transport aircraft and unique cargo ships). These transportation means are costly; monetary but also with regards to other resources such as lead time and CO2 emissions.

An optimal industrial system would not only minimise these KPIs but also ensure that all other constraints are kept in check. Such as the availability of suppliers for special parts, sourcing the same part from different suppliers (so called double sourcing) to increase the systems resilience or keep boundaries for the allotted workshare of suppliers in check to minimize overdependencies.

Optimizing such complex system topologies manually traditionally relies heavily on heuristics (trial and error, know-how, common sense) which usually fall short of achieving an optimal solution~\citep{aickelin2011heuristic}. However, due to the high dimensional complexity of the combinatorial problem, brute forcing an optimal solution is computationally expensive~\citep{karp1975computational}.

In order to provide a solid baseline for the exploration of this optimization problem, an industrial system data model was developed~\citep{Dietz_2023}. The model was built using Web Ontology Language (OWL), which was identified as a suitable machine-readable format for various industrial problems, from zero-defect manufacturing to supply chain management~\citep{ZHAO2008580,PSAROMMATIS2023103832,SCHEUERMANN2014913}. The model contains all possibilities for valid variants of the industrial system such as the product breakdown structure, production locations, transportation means, workshare constraints and valid network routes. This was done so that inconsistencies could be rectified more easily, efficiently share data with the optimization algorithm and - most importantly - to reduce the number of combinations only providing valid options to the optimizer.

This paper focuses on the optimization of the industrial system architecture, in particular the topology of production locations, sourcing strategies, and transport connections within these production networks.
The optimization process is split into two phases. The first phase, \phaseI, uses an evolutionary algorithm (EA) to create an optimized industrial system topology based on predefined constraints.
The second phase, \phaseII, represents transportation optimization together with batching optimization.

To verify the performance of the proposed optimization framework, real data from a specific case study of industrial system architecture design, provided by an involved aircraft manufacturer, was used. To the best of our knowledge, no study addressing a similar problem of this complexity has yet been described in the literature.

The main contributions of this paper are:
\begin{itemize}
    \item A new complex optimization problem for automatically designing industrial system architectures has been formulated. It involves challenges that every manufacturing company with a large network of manufacturing facilities and suppliers scattered across different countries faces today.
    \item An approach combining an evolutionary algorithm, formal mathematical programming methods, and an ontology-based data model was designed to solve the proposed optimization problem.
    \item Exploitation of OWL ontology to keep the knowledge base consistent, providing only relevant resources according to the ontology axioms.
\end{itemize}

This paper is organized as follows: First, a detailed problem description is provided in Section~\ref{sec:problem_definition}. 
Next, the main concepts of the proposed method are described in Section~\ref{sec:method}. 
Section~\ref{sec:related_work} presents a survey of related work.
Experiments and their discussion are presented in Section~\ref{sec:experiments}. 
Finaly, Section~\ref{sec:conclusions} concludes the paper.

\section{Problem Definition}
\label{sec:problem_definition}

This section provides a formal definition of the problem addressed in this work. 
The problem involves the design of an optimal industrial system architecture with respect to the overall transportation costs, subject to a number of constraints.

\begin{figure}[h]
    \centering    
    \includegraphics[width= 0.9\linewidth]{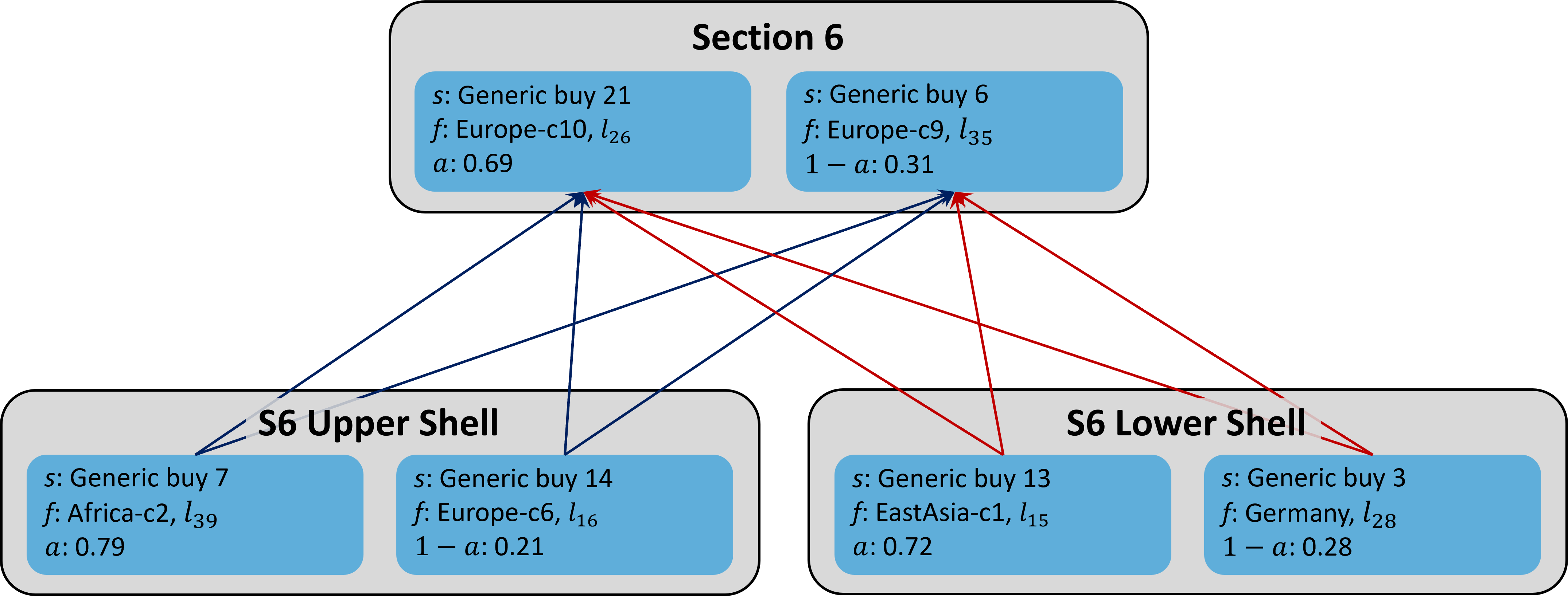}
    \caption{An illustrative example of a part of the production assignment realizing the production of \textbf{Section\,6} from two source parts, \textbf{S6 Lower Shell} and \textbf{S6 Upper Shell}, where the double sourcing is required.}
    \label{fig:double_sourcing}
\end{figure}

Following is a list of entities that appear in the problem definition:
\begin{itemize}
    \item Production graph \prodgraph\ -- the manufacturing process of the final product can be formally described by a tree, where the root node represents the \textit{final product}, the leaf nodes are atomic, further indivisible, parts of the product, and the internal nodes represent composite parts assembled from the parts that are at the input to the internal node.
    \item \parts\ -- a set of $M$ parts, $\{p_1, \dots, p_M\}$, present in the production graph \prodgraph, i.e., the final product, all composite parts, and all atomic parts.
    \item \psites\ -- a set of production site locations; in each location, one or more production plants are operated.
    \item \countries\ -- a set of countries where the production sites are located.
    \item \pplants\ -- a set of production plants (factories), where each production plant is assigned to a particular production site $l\in$\,\psites\ in one specific country $c\in$\,\countries. 
    Each plant can produce one or more types of parts. The set of parts producible by the plant $f$ is denoted as $P^f \subset$\,\parts.
    \item \suppliers\ -- a set of suppliers that can produce all elements of graph \prodgraph. 
    A supplier operates production plants at several production sites (one plant at a site) in various countries.
    \item \punits\ -- a set of production units \punits\, where each production unit is a unique pair of a particular supplier $s\in$\,\suppliers\ and one of its production plants $f\in$\,\pplants.
    \item \punitsp\ --  a subset of production units \punitsp\,$\subset$\,\punits\ such that each $u\in$\,\punitsp\ can produce the particular part $p\in$\,\parts.
    \item \punitss\ --  a subset of production units \punitss\,$\subset$\,\punits\ that belong to the supplier $s\in$\,\suppliers.
    \item \punitsc\ --  a subset of production units \punitsc\,$\subset$\,\punits\ that are located in the country $c\in$\,\countries.
    \item \warehouses\ -- a set of warehouses that serve as storage for input/output parts for/from several production plants. 
    \item \warehousesu\ -- a subset of warehouses \warehousesu\,$\subset$\,\warehouses\ that are available to the production unit $u\in$\,\punits. 
    \item \va\ -- a value-added \va\ is a real number from the interval (0, 1) that represents the ratio of the value of the part $p\in$\,\parts\ to the total value of the final product.
    \item \vac\ -- a value-added \vac\ is a real number from the interval (0, 1) that represents the actual value-added allocated to the country $c\in$\,\countries\ calculated as the ratio of the sum of values of all parts produced in production units $u\in$\,\punitsc\ to the total value of the final product.
    \item \vas\ -- a value-added \vas\ is a real number from the interval (0, 1) that represents the actual value-added allocated to the supplier $s\in$\,\suppliers\ calculated as the ratio of the sum of values of all parts produced in production units $u\in$\,\punitss\ to the total value of the final product.
    \item \vacmin, \vacmax\ -- required minimum and maximum boundaries on the value-added allocated to the country $c\in$\,\countries. 
    \item \vasmin, \vasmax\ -- required minimum and maximum boundaries on the value-added allocated to the supplier $s\in$\,\suppliers. 
    \item \vaumax\ -- the maximum added-value allowed for the production unit $u\in$\,\punits.   

    \item \trmeans\ -- a set of possible means of transportation that can be used to transport the parts. Each type of transportation $t\in$\,\trmeans\ has specified its CO$_2$ emission \tremission, speed \trspeed, cost \trcost, and maximum load volume \trload. 

    \item \trmgraph$(\mathcal{N},\mathcal{E})$ -- transportation graph is a directed graph in which all possible instances of the industrial system can be represented. Its nodes, $\mathcal{N}$, comprise all production units and all available warehouses, i.e., $\mathcal{N}=U \cup W$. 

    The edges in the set $\mathcal{E}$ represent transportation links used to carry a load from the source node to the destination node, i.e., an edge is an oriented link $(u,v)$ where $u,v \in \mathcal{N}$.
    Each edge $\varepsilon \in \mathcal{E}$ can only choose its transportation type $\tau^\varepsilon$ from the set of possible transportation types $T^{\varepsilon} \subset T$ available for the edge. 
    The chosen transportation type determines the edge's length $l^\varepsilon$.
    Further, each edge has its batch size $b^\varepsilon$, which is the number of parts transported in a single shipment along the edge.
    Here, we consider a single-product batching consisting of one or more pieces of the same part. Figure~\ref{fig:transportLinksDist} illustrates the most important components of the transportation graph.

    \item \indsys$(N, E)$ -- a directed graph as a sub-graph of the transportation graph representing a particular industrial system designed to produce the final product. Its nodes $N$ and edges $E$ are a subset of $\mathcal{N}$ and $\mathcal{E}$, respectively.
    
    Every shortest path in terms of the number of edges between two nodes $u_1,u_2\in$\,\punits\ must have the following structure $u_1 \rightarrow w^* \rightarrow u_2$, where $w^*$ means that there can be zero or more warehouses $w\in W$ used on the path. Each edge $e\in E$ has already assigned its particular type of transportation that determines its length $l^e$. It has also specified its batch size $b^e$.
    Each production unit node $u \in N \cap U$ is assigned one or more specific parts chosen from its production plant's set of producible parts. In Figure~\ref{fig:double_sourcing}, the blue tiles represent production units with its supplier \textit{s} and production plant \textit{f}.
     
    In the case of \textit{double sourcing}, each part has to be produced in two different production units while splitting the whole production of the part between the two production units in the ratio $a$~:~$(1-a)$ where $0 < a < 1$, see Figure~\ref{fig:double_sourcing}. The particular value of the production split is defined for all parts and respective production units of the industrial system.
    
    The assignment of parts to production units and the production split definitions are denoted as a \textit{production assignment} of the industrial system.
\end{itemize}

\subsection{Constraints}
\label{sec:constraints}

The constraints imposed on a valid solution used in this study are defined below. 
\begin{itemize}
    \item \textbf{Constraint\,1} -- a part shall only be manufactured by a production unit with the manufacturing capability for said part.
    \item \textbf{Constraint\,2} -- in the case of double sourcing, the two production units producing the same part shall be located in two different countries if applicable. Note that some parts can be produced only in production units where all of them are located in the same country. In such a case, this constraint is not enforced.
    \item \textbf{Constraint\,3} -- in the case of double sourcing, the split between production units producing the same part shall be between 20\% to 80\%, i.e., $0.2 \leq a \leq 0.8$. 
    \item \textbf{Constraint\,4} -- the value added by the country $c$ is less than or equal to \vacmax\ for all $c\in$\,\countries.
    \item \textbf{Constraint\,5} -- the value added by the supplier $s$ is less than or equal to \vasmax\ for all $s\in$\,\suppliers.
    \item \textbf{Constraint\,6} -- the value added by the production unit $u$ is less than or equal to \vaumax\ for all $u\in$\,\punits.
    \item \textbf{Constraint\,7} -- the batch size $b^e$ assigned to every edge $e$ of the industrial system \indsys$(N,E)$ can not be larger than the maximum possible value, which is determined by the maximum load's volume \trload\ of the transportation type assigned to the edge and the size of the part transported along the edge.
\end{itemize}

Some constraints are relevant only for \phaseI\ and \phaseII, respectively; see below in Section~\ref{sec:phaseI} and Section~\ref{sec:phaseII}.

\subsection{Optimization objectives}
\label{sec:objectives}

When designing the optimal industrial system, we consider the following optimization objectives:
\begin{itemize}
    \item Minimize the total amount of CO$_2$ generated while using the industrial system to produce $K$ pieces of the final product. 
    Formally, this objective is defined as
    \begin{equation*}
        \operatorname*{arg\,min}_{I(E,N) \in \mathcal{G^T(E,N)}} f_1(I(E,N))=\sum_{e\in E} n_e * l^e * \tau^e_c
    \end{equation*}
    where $n_e$ is the number of rides carried out along the edge $e$ of the length $l^e$ while using its assigned transportation type $\tau^e$ and emitting $\tau^e_c$ grams of CO$_2$ per kilometer.
    
    \item Minimize the total transportation time spent while using the industrial system to produce $K$ pieces of the final product.
    Formally, this objective is defined as
    \begin{equation*}
        \operatorname*{arg\,min}_{I(E,N) \in \mathcal{G^T(E,N)}} f_2(I(E,N))=\sum_{e\in E} n_e * \frac{l^e}{\tau^e_s}
    \end{equation*}
    where $n_e$ is the number of rides carried out along the edge $e$ of the length $l^e$ km using the transportation type $\tau^e$ with the speed of $\tau^e_s$ km/hour.

    \item Minimize the total transportation distance while using the industrial system to produce $K$ pieces of the final product.
    Formally, this objective is defined as
    \begin{equation*}
        \operatorname*{arg\,min}_{I(E,N) \in \mathcal{G^T(E,N)}} f_3(I(E,N))=\sum_{e\in E} n_e * l^e
    \end{equation*}
    where $n_e$ is the number of rides carried out along the edge $e$ of the length $l^e$ km using the transportation type $\tau^e$.

    \item Minimize the total transportation costs while using the industrial system to produce $K$ pieces of the final product.
    Formally, this objective is defined as
    \begin{equation*}
        \operatorname*{arg\,min}_{I(E,N) \in \mathcal{G^T(E,N)}} f_4(I(E,N))=\sum_{e\in E} n_e * l^e * \tau^e_t
    \end{equation*}
    where $n_e$ is the number of rides carried out along the edge $e$ of the length $l^e$ km using the transportation type $\tau^e$ with the transportation cost of $\tau^e_t$ euro/km.
\end{itemize}





\section{Method}
\label{sec:method}

\subsection{Knowledge Base}

This section introduces the benefits of the knowledge base which is exploited by {\phaseI} and {\phaseII}. Detailed information about the knowledge base is described in~\citep{Dietz_2023} as mentioned in the introduction section.

Input data are specified within Web Ontology Language (OWL) ontology~\citep{antoniou2009web}. This formalism offers the possibility to utilize not only data but also knowledge as machine-readable. The machine readability denotes mainly semantics added into the dataset, i.e., specified relations between entities, cardinality of properties, and other axioms in general.

The important benefits of OWL ontologies are the exploitation of Semantic Web Rule Language (SWRL)~\citep{horrocks2004swrl} and automated reasoning~\citep{shearer2008hermit}. SWRL to move from Open World Assumption~\citep{drummond2006open,WANG20141041} paradigm of OWL to the closed world. Furthermore, applications for this ontology are described in~\citep{Dietz_2023}. Next, reasoning tasks are the following:
\begin{itemize}
    \item 
    Satisfiability of a concept - if the concept definition is not contradictory.
    \item
    Subsumption of concepts - if concept C subsumes concept D.
    \item
    Consistency of ABox with respect to TBox~\citep{de1996tbox} - if individuals in ABox do not violate axioms defined by TBox.
    \item
    Check an individual - if an individual is an instance of a concept.
    \item
    Retrieval of individuals - all individuals of a concept.
    \item
    Realization of an individual - all concepts specifying the individual.
\end{itemize}

In this work, the consistency check reasoning task is exploited. Not only to maintain data to be in required shape, but also to check if a product produced by a production unit has an appropriate transportation resource. I.e., whether the bounding box of a part fits into a container.

\subsection{Phase I}
\label{sec:phaseI}

This section describes the \phaseI\ part of the proposed method, i.e., the EA designed to search for the best production assignment of the industrial system. 
We chose an evolutionary algorithm because these algorithms have proven to be highly suitable for solving complex real-world constrained optimization problems; see \cite{PARK201668, CHAI2025112732, LI2025112650}.
First, the optimization objectives and constraints relevant to this phase are listed. Then, the EA's main components and working schema are described.

\subsubsection{Optimization objectives and constraints}
\label{sec:phaseI_objectives_constraints}

In this phase, out of all constraints defined in Section~\ref{sec:constraints}, only \text{constraints\,1--\,6} are considered. Constraint 7 is irrelevant since only the production assignment is optimized here. The optimization of transportation utilizing the information about available transportation means, possible batching strategies, takt time, the required number of products to be produced per month, etc., is left for the \phaseII.

The EA uses two performance measures listed below, ranked by priority:
\begin{itemize}
    \item Primary objective -- maximize the satisfaction ratio, \sr, defined as the ratio of the number of parts successfully added to the production assignment to the total number of parts. Successful insertion of parts into the production assignment means that the parts were added to the production assignment without violating any of the relevant constraints; see below in Section~\ref{sec:ea}.
    \item Secondary objective -- minimize the overall transportation distance, \dist, defined as the sum of distances between all pairs of the source and destination production units involved in the production assignment. For a particular edge $\varepsilon=(u_1, u_2)$, where $\varepsilon \in \mathcal{E}$ and $u_1, u_2 \in U$, the maximum distance over all possible distances defined by the set of transportation types available for the edge, $\tau^\varepsilon \in T^\varepsilon$, is considered. This upper bound value is used since it is unknown at this stage what transportation means will be used to transport products from the unit $u_1$ to the unit $u_2$.
    Thus, the measure serves as a rough estimate of transportation costs. It is calculated only for a feasible solution with \sr\,$= 1$.
\end{itemize}

\subsubsection{Evolutionary Algorithm}
\label{sec:ea}

This section describes the main components of the EA designed to generate a valid production assignment of the industrial system.

\textbf{Solution representation}. 
The EA implemented in this work adopts the indirect representation of solutions introduced in \citep{Kubalik19}. In this representation, the exact assignment of parts to production units is not explicitly encoded in the solution representation. Instead, the solution is represented using a list of parts, denoted as \textit{priority list}, which determines the order in which individual parts will be inserted into the production assignment. 

When evaluating the quality of a particular candidate solution, its priority list must be first translated into the production assignment via an iterative mapping process. It starts with an empty production assignment to which new parts are gradually added one by one in the order defined by the priority list.
When adding a new part $p$ to the current production assignment, the set of available production units is first initialized, starting with the set of all production units that can produce the part, $U^p$, and filtering out those units that already violate some constraint.
If the set of available production units is not empty, one of them is randomly chosen, preferably among the unused ones (i.e., those not assigned to produce any part yet). Otherwise, no production unit can be assigned to the part, and the mapping process terminates.
Thus, the mapping procedure ends when all parts have been successfully incorporated into the production assignment or when a part that cannot be added to the production assignment has been reached. The former means that a complete and valid production assignment has been constructed. The latter means that the priority list does not represent a valid solution.

\textbf{Genetic operators}.
Here, we adopt a variation of the 1-point crossover operator. It randomly selects one crossover point, inherits the head section from the first parent's priority list, and fills in the remaining missing elements in the order they appear in the second parent's priority list. The mutation operator operates on a single parent's priority list so that it changes the position of a randomly chosen part within the priority list.

A standard tournament selection is used, which applies the following rule to decide if the solution $s_1$ is better than the solution $s_2$:
\begin{align*}
\begin{split}
 \textbf{if}\text{  } & (\text{\sr}(s_1) > \text{\sr}(s_2))  \textbf{  or}\\
                & (\text{\sr}(s_1) = \text{\sr}(s_2) \text{  \textbf{and}  } \text{\dist}(s_1) \leq \text{\dist}(s_2)) \\
 \textbf{then}\text{  } & \text{return true}
\end{split}
\end{align*}
This rule puts pressure towards production assignments, which are valid and induce a small overall transportation distance. 
Note, the \dist\ metric does not consider possibilities to transport products via warehouses. This is left for the second optimization phase.
It is important to note that the validity of the solution, whether complete or incomplete, is ensured by the proposed method for mapping the priority list to the production assignment. Convergence to complete valid solutions is driven by the selection rule.

\textbf{Evolutionary model}.
The algorithm evolves a population of candidate solutions using the selection, crossover, and mutation operators described above. 
The outline of the algorithm is shown in Algorithm~\ref{alg:ea}. It starts with random initialization and evaluation of the population of solutions, \pop. Specifically, a population of solutions is generated so that each solution is assigned a randomly generated priority list as a random permutation of all parts, which is then mapped into the corresponding production assignment and its performance parameters \sr\ and \dist\ are calculated.
The algorithm then iterates through a specified number of generations, lines~\ref{alg3}--\ref{alg7}. In each generation, an intermediate population of solutions, \interpop, is created using the solutions of the current population \pop, lines~\ref{alg4}--\ref{alg5}. First, parental solutions are selected, which then undergo the crossover and mutation operations to create their offspring solution \child. Each newly created \child\ solution is evaluated.
Once the \interpop has been completed, it is merged with the current \pop, resulting in a new version of the \pop (line~\ref{alg6}). While merging the two populations, only unique solutions are kept. The better solutions are preferred and ties are broken in favour of the new \interpop solutions. 
Finally, $n$ best solutions of the final population are returned as the output of the run. 
%
\LinesNumbered
\DontPrintSemicolon
\begin{algorithm}[!t]
\linespread{1.2}\selectfont
\fontsize{8.6}{9.6}\selectfont
\vspace{-0.05em}
   \gen $\leftarrow 0$ \;
   \pop.\init{}\;\nllabel{alg1}
   \While{\gen $<$ \maxgens} { \nllabel{alg3}
      \gen $\leftarrow$ \gen + 1 \;
      \interpop $\leftarrow$ \texttt{\{\}} \;
      \While{\interpop.\size{} $<$ \pop.\size{}} {\nllabel{alg4}
         \parentI $\leftarrow$ \pop.\selectT{} \;
         \parentII $\leftarrow$ \pop.\selectT{} \;
         \If{\rand{} $<$ $p_c$}{
             \child $\leftarrow$ \parentI.\cross{\parentII} \;
             \If{\rand{} $<$ $p_m$}{
                 \child $\leftarrow$ \child.\mutate{} \;
             }
         }
         \Else{
             \child $\leftarrow$ \parentI.\mutate{} \;
         }
         \child.\eval{}  \;
         \interpop.\add{\child}
      }   \nllabel{alg5}
      \pop $\leftarrow$ \merge{\pop, \interpop} \;\nllabel{alg6}
   } \nllabel{alg7}
\KwRet{\pop.\getbest{$n$}} \;
\caption{Evolutionary algorithm}
\label{alg:ea}
\end{algorithm}

\subsection{Phase II}
\label{sec:phaseII}

Upon completion of \phaseI\ and generating the production assignment, the subsequent \phaseII\ resides in completing the industrial system with transportation links between industrial system components, with the quantity of required parts production based on site shares, with information about production duration, and with transportation details --- duration, costs, $CO_2$ emission, and how many batches are required to satisfy demand in target sites.

Thanks to EA multicriteria optimization, which considers not only the solution's satisfiability but also minimize the transportation distance, the space of all possible solutions is significantly pruned. Thus, the task of this phase represents forming the graph of sites and possible connections between sites (together with detailed transportation information), supplemented by warehouses. The output of the first phase includes not only information about producers and consumers, but also information about production shares. Therefore, this task may be solved by leveraging local optimization propagating through the graph against production time. 

\subsubsection{Time Matters}\label{sec:timeMatters}

To bring the solution closer to reality, it is important to take into account the time variable. Compared with other approaches, for example, simulations in which time may be treated in a continuous (or discrete) infinite manner, the proposed approach in this solution restricts the duration of the overall process to an interval according to the amount demanded of the final product (i.e., the number of ``Single Aisle Aircraft'').

The idea of how to cope with the time issue is as follows. First, the time interval is computed for the final assembly and, as the next step(s), it is spread through the industrial system to the lowest levels where fundamental parts are produced. Batching is taken into account in this step to find out how many turns/deliveries are required for every transportation link. Dependencies on more various input parts for many production sites may lead to delays which are covered by this approach.

Furthermore, the solution more precisely approximates the real-life situation in terms of time and logistics with an increasing number of final products.



\subsubsection{Transportation Network Generation}

First, a transportation network has to be generated. The generation process is based on input from the first phase, that is, the already defined architecture of the industrial system in the form of the production assignment. 
Thus, the backbone of the producer/consumer network is available and has to be extended by the following components and properties:
\begin{itemize}
    \item 
    \textbf{Edges between nodes}, where nodes represent production sites (i.e., producers and consumers). Every edge represents a specific type of transportation with relevant properties --- the number of given products in one container, the number of containers required for demand satisfaction, transportation time, distance, transportation cost, and $CO_2$ emissions.
    \item 
    \textbf{Warehouse nodes} and their connections. The involvement of warehouses in transportation network represents a complex task. This approach relies on flows between producers and consumers already defined by the evolutionary algorithm (from the first phase), and, therefore, the task is to decide if the material is transported directly or routed via one or more intervening warehouses (for example, source or/and target warehouses where source warehouse is located nearby producers and target warehouse is located nearby consumer/consumers). The ``nearby'' property of warehouses refers to ``isNearby'' OWL object property from the source ontology. A mixed batching strategy is taken into account for links between warehouses in the case where the various parts are available in the warehouse and are required to be transported together at the same time.
\end{itemize}

\begin{figure*}[t!h!]
    \centering
    \includegraphics[width=0.9\linewidth]{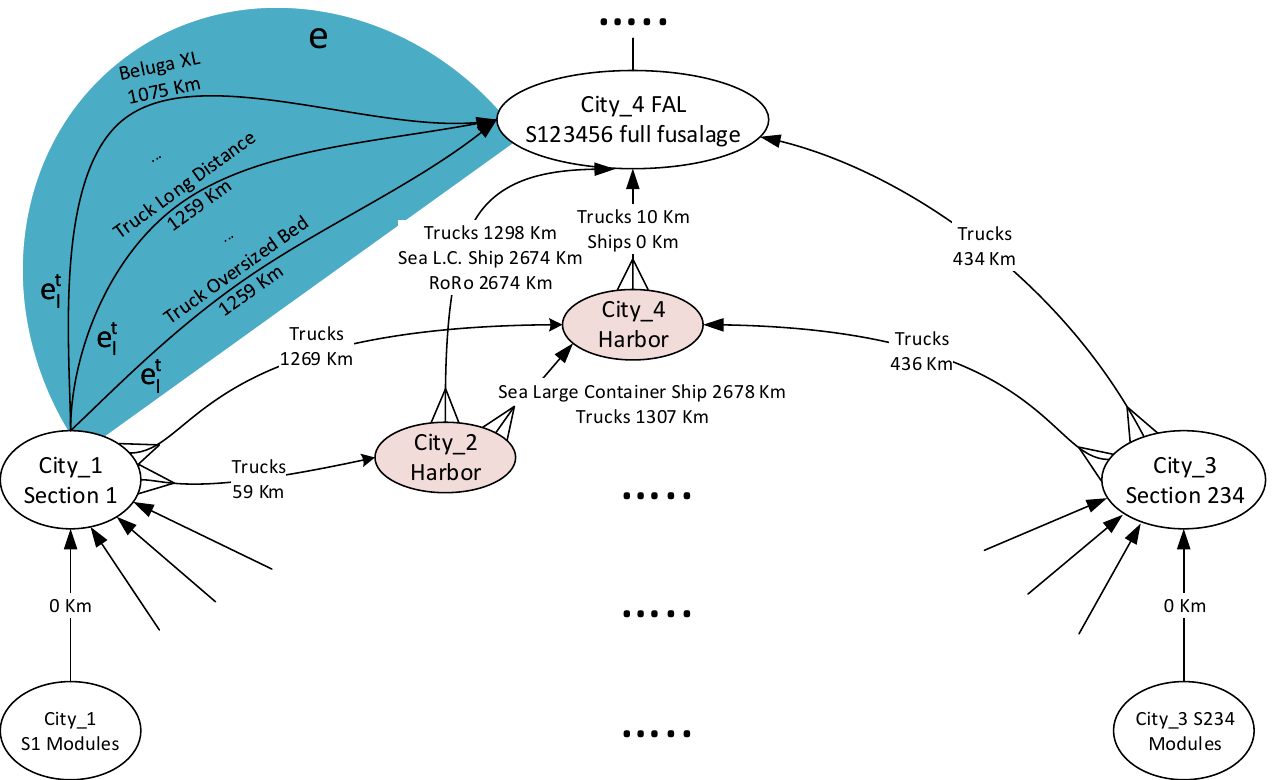}
    \caption{Part of the transport network with transport links, distances according to transportation means, and two warehouses. In the depicted oriented graph, vertices represent production units and edges between vertices represent transport links between production units. There is typically more than one edge between two vertices, where every edge stands for a different transportation resource. There are two warehouses named ``City\_2 Harbor'' and ``City\_4 Harbor''.}
    \label{fig:transportLinksDist}
\end{figure*}


The outcome of the transportation network generation task is the directed acyclic graph with possibly redundant edges between two nodes in the case of more possible transportation means. The part of the transport network with transport links together with distances by possible transportation means is shown in Fig~\ref{fig:transportLinksDist}. The last thing to be specified in this step is the specification of the required amount of parts and their type/s. Together with the related computation of the batching task (described in the following section), it is done in a top-down manner.

\subsubsection{Batching Strategies}

Batching solves the important task of how to load parts into a cargo container/mean. The inputs are the type of part (resp. parts) to be delivered, its (resp. their) dimensions, and the required amount. Afterwards, the output is information on how many parts fit into the container or transportation means and how many deliveries are required to fulfil the given demand.


\begin{itemize}
    \item 
    Single-batching strategy --- items to pack are of one particular type. Primarily, it represents a batching for transportation for direct transportation links, that is, directly between a source and a target manufacturing resource (site).
    \item 
    Mixed-batching strategy --- items to pack are represented by various part types. It represents a batching between two warehouses, where the items may be mixed. 
\end{itemize}

Two bin-packing algorithms were tested --- Largest Area Fit First algorithm~\citep{gurbuz2009efficient} and the GRASP heuristics-based algorithm (swap and relocate neighbourhood search)~\citep{layeb2012novel}. The precise shape of parts is not considered, and every part is represented by its bounding box. For this application, both of the aforementioned algorithms have comparable results. The GRASP algorithm was selected for use in this solution. The sample output of the batching task is presented in the following listings.

\begin{lstlisting}[basicstyle=\small, caption={An output from the batching algorithm describing bounding box size of the part, size of the container, demanded amount of the part, number of containers (resp. number of deliveries), part type, and type of the transportation resource.}]
part size:          [12300, 2300, 1800]
container size:     [14800, 3300, 3000]
demand:             3
nb of containers:   3
part:               Horizontal Tailplane
transportation:     Truck Oversized Low Bed
\end{lstlisting}

\begin{lstlisting}[basicstyle=\small]
part size:          [6800, 400, 1500]
container size:     [2330, 11998, 2350]
demand:             8
nb of containers:   2
part:               Pylon Right
transportation:     Sea Large Container Ship
\end{lstlisting}
With the help of batching, we have information about not only how many parts fit into a given cargo capacity but also how invalid transport links may be pruned. In other words, if the part cannot be loaded into a container, then the transport link between the source site and the target site using a certain transportation means is considered invalid and is not taken into consideration for further processing.

In the case of a shared part of the transport link between the source production unit, one or more warehouses, and the target production unit, the mixed batching is applied for the shared and continuous path. Finally, information about batches (single or mixed) is stored in the transportation network graph to the relevant transport links.

\subsubsection{Transport Optimization}

The pruned space of possible solutions from the result of the \phaseI is used for the final optimization. As mentioned in previous paragraphs, possible solutions form the graph, which contains information about $CO_2$, costs and duration within the supply chain, distance, batch size, and batch count with respect to the available transportation means.

Based on the nature of the task and its limited scale/size after the preceding processing, local optimization propagating through the graph against production time was proposed as sufficient and suitable. The Descent Recursive Acyclic Graph Optimizer (DRAGO) algorithm aims to optimize transportation between production units and warehouses. The pseudo-code is presented in Algorithm~\ref{alg:drago}. Given a starting vertex (i.e., final assembly line --- line 2 in Algorithm~\ref{alg:drago}) and an optimization criterion (e.g., duration, cost, emissions), DRAGO recursively explores the graph in a top-down manner. This is achieved by the \texttt{OptimizeInput} function (line 10), which, for a given vertex and input part, evaluates all relevant incoming transportation links based on the defined optimization criteria. The relevant incoming transportation links are selected by \texttt{AdjacentEdges} function (line 14), which finds input links to vertex transporting specified input product. Next, the link with the best score is selected. If the selected transportation link involves a mixed batch (indicated by transport\_link.parts.size $>$ 1 --- line 19), the current vertex-input pair is added to a mixed\_batching\_inputs list (line 24) for subsequent joint processing. Otherwise, the selected best\_transportation\_link is added to the optimal\_path (line 26). The algorithm then recursively calls \texttt{OptimizeInput} for the source vertex of the chosen transportation link and each of its required input parts (line 29), effectively propagating the optimization down the supply chain.

\begin{algorithm}
\linespread{1.2}\selectfont
\fontsize{8.6}{9.6}\selectfont
\vspace{-0.05em}
\caption{DRAGO}
\label{alg:drago}
\SetKwFunction{FDrago}{Drago}
\SetKwFunction{FOptimizeInput}{OptimizeInput}
\SetKwFunction{FAdjacentEdges}{AdjacentEdges}
\SetKwProg{Fn}{Function}{:}{}
\CommentSty{REQUIRE Acyclic graph $G(V, E)$, where $V$ are production units or warehouses and $E$ are transportation links with properties (transportation mean, distance, $CO_2$ emissions, duration, size of batch, number of drives, transportation costs)}\;
\CommentSty{REQUIRE Starting vertex $v_{start}$ (final assembly line)}\;
\CommentSty{REQUIRE Optimization criteria (e.g., duration, distance, $CO_2$ emissions, transportation costs)}\;
$optimal\_path \leftarrow []$\;
$mixed\_batching\_inputs \gets []$\;
\Fn{\FDrago{}}{
\For{each $input\_part \in v_{start}$}{
    \FOptimizeInput{$v_{start},input\_part$}\;
}
\KwRet{$optimal\_path$}
}
\Fn{\FOptimizeInput{$vertex, input$}}{
    $best\_transportation\_link \gets \text{NULL}$\;
    $best\_score \leftarrow \infty$\;
    $mixed\_batch \gets \text{FALSE}$\;
    \For{each $transportation\_link \in$  \FAdjacentEdges{$vertex,input$}}{
        Calculate $score$ based on optimization criteria\;
        \If{$score < best\_score$}{
            $best\_transportation\_link \leftarrow transportation\_link$\;
            $best\_score \leftarrow score$\;
            \If{$transport\_link.parts.size > 1$}{
                $mixed\_batch \gets \text{TRUE}$\;
                \textbf{break}\;
            }
        }
    }
    \If{$best\_transportation\_link \neq \text{NULL}$}{
        \If{$mixed\_batch$}{
            Append ($vertex,input$) to $mixed\_batching\_inputs$
        }
        \Else{
            Append $best\_transportation\_link$ to $optimal\_path$\;
            }
        $current\_vertex \leftarrow \text{source vertex of } best\_transportation\_link$\;
        \For{each $input\_part \in current\_vertex$}{           \FOptimizeInput{$current\_vertex,input\_part$}\;
        }
    }
    \KwRet\;
}
\end{algorithm}

\section{Related work}
\label{sec:related_work}

In this section, we provide an overview of recent works dealing with the design of distributed manufacturing and the related optimization of the transportation network. Note that our approach addresses all these aspects jointly, including additional factors such as batching and the use of distributed buffer warehouses.


Research on multi-site production planning has addressed how to assign the manufacturing of various parts or products to a network of available plants. For example, Shen et al. \citep{rw01-SHEN2020451} develop a multi-plant production planning model that incorporates practical features like non-repeated setup operations and aperiodic shipment schedules. They formulate the problem as a mixed-integer linear program (MILP) with additional linearized constraints, solving it to optimality to maximize total profit under these complex setup and shipment conditions. In a similar vein, Çömez-Dolgan et al. \citep{rw02-COMEZDOLGAN20231033} consider a manufacturer with multiple plants facing regional demands for a wide product assortment. Their work focuses on deciding which products to produce at each plant (assortment planning) when trans-shipments between plants are allowed at an extra cost. The authors propose an optimization model to maximize profit, and derive structural properties (such as nested optimal assortments) of the solution.
Han et al. \citep{rw03-HAN20191} address a two-level supply chain where multiple suppliers produce semi-finished parts that are sent to a central assembly plant. They combine a cost-minimization MILP to select suppliers and production quantities each period with a tailored heuristic algorithm. Similarly, Wibawa et al. \citep{rw04-Wibawa2022} explore metaheuristic approaches for multi-site aggregate production planning. They consider a textile manufacturer with several production sites and formulate a multi-objective aggregate planning problem to minimize costs. They employ a particle swarm optimization (PSO) algorithm to search for near-optimal production allocation plans. Their PSO-based method demonstrated in a real case study that it can effectively handle large-scale multi-factory allocation problems.


Another stream of related work focuses on optimizing transportation and distribution networks in global supply chains. These studies typically assume production locations are given and seek to minimize logistics costs by better routing, scheduling, and network design. Guo et al. \citep{rw05-Guo2022} investigate a spare-parts distribution network that must supply manufacturing plants over multiple periods. They present a dynamic nonlinear programming model that jointly manages inventory at customer facilities and flow decisions in the spare-parts network. To solve this complex problem, the authors develop an improved self-adaptive dynamic PSO algorithm that quickly adapts to changing conditions.
Peng et al. \citep{rw06-PENG2022781} study an integrated transportation planning problem in a retail supply network under carbon emission regulations and demand uncertainty. They use uncertainty theory to model ambiguous data (e.g., demand fluctuations) and formulate four mathematical programming models corresponding to different carbon regulatory policies. 
In the domain of inbound logistics, Quan et al. \citep{rw07-Quan2021} focus on optimizing milk-run routes for delivering auto parts from multiple suppliers to an assembly plant under low-carbon objectives. They extend the classic vehicle routing problem by adding fixed delivery schedules (time windows), fuel consumption, and carbon emission costs into the routing model. An improved ant colony optimization algorithm is then applied to find efficient pickup routes. 
Transportation network optimization can also be integrated with inventory decisions. Vicente \citep{rw08-Vicente2025} proposes a MILP-based planning tool that coordinates inventory control and shipment scheduling across a multi-echelon distribution system. The model determines optimal replenishment quantities and shipping plans for warehouses and retailers, following a periodic-review inventory policy. 
Experiments with realistic supply chain data show that jointly optimizing inventory and transportation (rather than treating them separately) can significantly lower overall supply chain costs.


Integrated approaches consider production allocation and transportation planning decisions simultaneously and capture trade-offs between manufacturing and logistics. Klenk et al. \citep{rw09-Klenk2022} present a comprehensive model for global production network configuration, which jointly optimizes product allocations to plants and the distribution network over a planning horizon. Their approach is multi-objective: it minimizes cost and other metrics while incorporating flexibility and reconfiguration considerations. Using a preemptive goal programming method, they generate Pareto-optimal solutions that balance conflicting objectives. 
Multi-objective mathematical programming is a common tool in integrated production-distribution planning. Badhotiya et al. \citep{rw10-Badhotiya2019} formulate a fuzzy multi-objective MILP for a two-echelon supply chain with multiple plants and distribution centers. Their model optimizes production quantities at each plant and product flows to distribution centers, aiming to maximize profit and service level while minimizing costs. They introduce fuzzy goal programming to handle the trade-offs among objectives and uncertainties in demand. 
The use of fuzzy objectives provides decision-makers flexibility in prioritizing goals in an integrated plan.
Another notable contribution is by Neiro et al. \citep{rw11-NEIRO2022107778}, who tackle a production-distribution coordination problem in the industrial gas industry. They consider a network of production plants producing liquid argon and a set of customers served by tanker trucks. The challenge is to balance production levels at each site with vehicle routes and delivery schedules to meet customer demand at minimum cost. The authors adopt a two-phase solution: first generate efficient delivery routes via a routing heuristic, and then include those routes in a unified MILP that optimizes both production schedules and distribution assignments. This decomposition approach yields high-quality integrated plans for a real-world scale problem. 
%
Recent studies have started to address integrated production–distribution planning with explicit sustainability and uncertainty considerations in industrial contexts. For example, Xue et al. \citep{rw12-Xue2023} develop a multi-objective model for joint production-distribution optimization in an automobile supply chain under sustainability constraints. Their approach considers a hyper-connected, Physical Internet-enabled order-to-delivery system, optimizing both operational cost and environmental performance measures. By applying an improved NSGA-III algorithm, this work demonstrates that substantial gains in efficiency and eco-performance can be achieved simultaneously in a realistic automotive scenario. 
\section{Experiments}
\label{sec:experiments}

In this section, we present proof-of-concept experiments to demonstrate the effectiveness of the proposed approach.
First, \phaseI\ experiments on two problem types, single and double sourcing, are presented.
Then, \phaseII\ experiments with the following five variants of the constraint optimization problem are described:
    \begin{enumerate}
        \item Find the industrial system $I(E,N)$ such that the objective $f_1(\cdot)$ is minimized.
        \item Find the industrial system $I(E,N)$ such that the objective $f_2(\cdot)$ is minimized.
        \item Find the industrial system $I(E,N)$ such that the objective $f_3(\cdot)$ is minimized.
        \item Find the industrial system $I(E,N)$ such that the objective $f_4(\cdot)$ is minimized.
        \item Find the industrial system $I(E,N)$ such that both objectives $f_1(\cdot)$ and $f_2(\cdot)$ are minimized simultaneously.
    \end{enumerate}

The first four variants are single-objective optimization problems, while the fifth one is a bi-objective optimization problem. 
In all five cases, the optimization task faces manifold challenges. First, a valid and well-performing production assignment must be generated in \phaseI. Then, given the production assignment found, finding the optimal transportation graph in \phaseII\ requires choosing its optimal topology (i.e., choosing between direct part transportation or routing via warehouse(s)) and optimally choosing the attributes of all transportation links, the type of transport and the size of the batch.

\subsection{Experiment setup}
\label{sec:exp1_setup}

\begin{figure*}[h!]
    \centering
    \includegraphics[angle=0, scale=0.38]{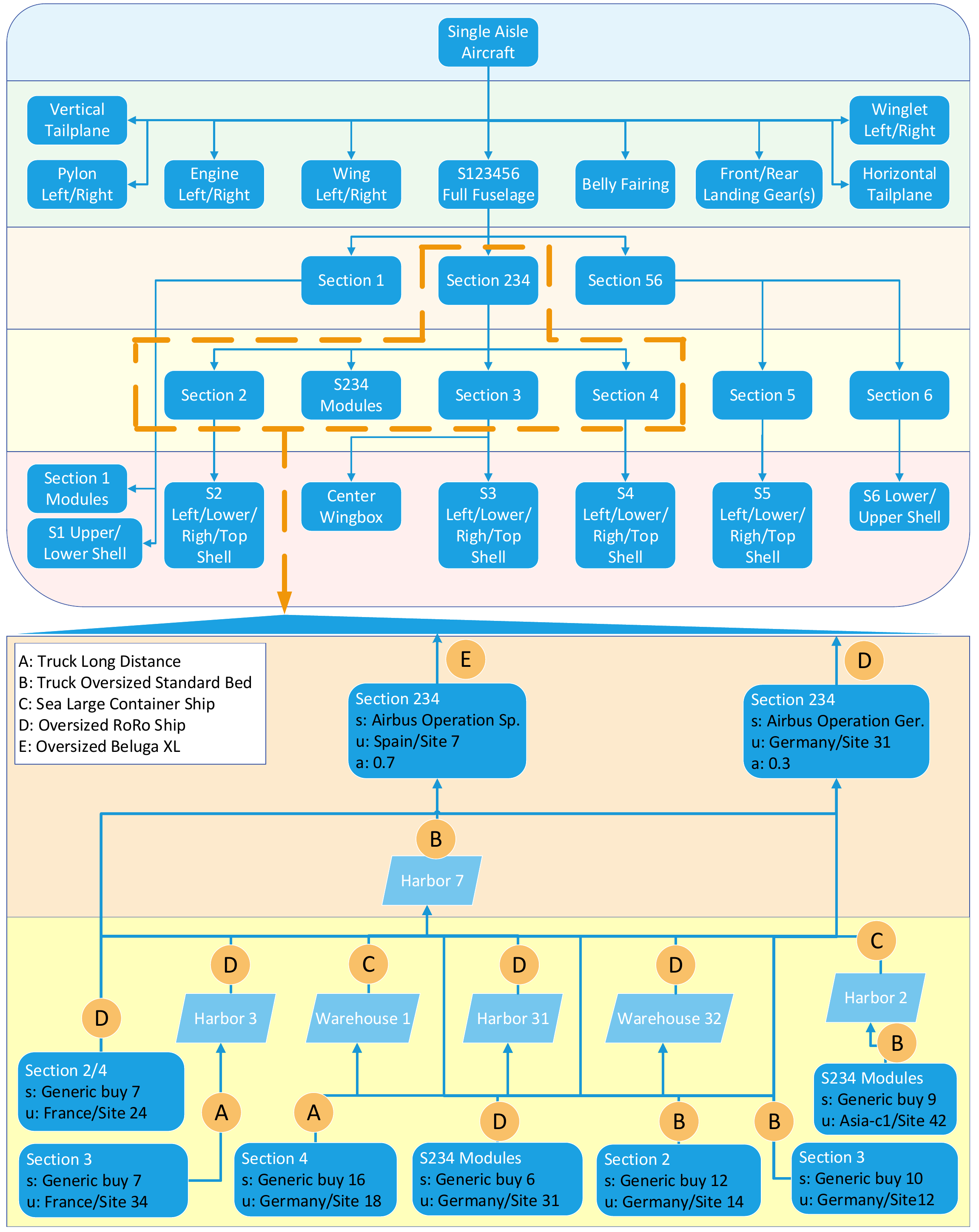}
\caption{The production graph (top) and an example of a part of the optimized industrial system with details of the corresponding transportation graph (bottom). Arrows in the production graph represent the specialisation relation.}
\label{fig:airbus_decomp-3}
\end{figure*}

To test our approach, we used a case study from the aerospace manufacturing sector with the following parameters: $|P|=M=47$, $|C|=17$, $|S|=29$, $|L|=43$, $|U|=45$, $|W|=34$, $|T|=17$ (9 types of maritime transport, 6 types of freight road transport, 2 types of air transport). Each transportation type has its CO2 emission, speed, and maximum load volume defined.
The upper part of Figure~\ref{fig:airbus_decomp-3} shows the respective production graph.
The required parameters of the industrial system sought are:
\begin{itemize}
    \item The maximum value added by all countries but France, USA and the United Kingdom is set to \vacmax\,=\,0.1, see Table~\ref{fig:exp1_countries}. France and the United Kingdom used \vacmax\ set to 0.22 and 0.12, respectively, due to the single-country parts produced only in these two countries.
    
    The USA used \vacmax\ set to 0.2 since it is one of the two producers (the other one is France) of ``engine right'' and ``engine left'', where both products have the value added of 0.1. Also, the USA has two final assembly line (FAL) production units, producing the ``Single Aisle Aircraft'' whose value added is 0.11. Thus, setting the \vacmax\ to the value 0.2 makes it possible to reasonably utilize the USA production units.
    
    \item The maximum value added by a supplier is the same \vasmax\,=\,0.15 for all suppliers.
    \item The maximum value added by a production unit is set to \vaumax\,=\,0.15 for all production units.
\end{itemize}


\subsection{Phase I - Experiments}
\label{sec:experiment1}

This section presents the empirical evaluation of the EA. The EA was run with the following configuration: population size 500, number of generations $G=200$, tournament size 3, crossover rate $p_c=0.8$, mutation rate $p_m=0.1$.

\subsubsection{Results}
\label{sec:exp1_results}

Table~\ref{tab:exp1_single} and Table~\ref{tab:exp1_double} show results obtained for the ``single sourcing'' and ``double sourcing'' scenario, respectively. The single sourcing scenario considers a single production unit assigned to all parts but the Single Aisle Aircraft\footnote{The number of parallel FAL production units required for the Single Aisle Aircraft can be chosen between 1-5 depending on the investigated scenario, i.e., depending on the envisioned throughput of the system.}, for which two production units are required.
The double sourcing scenario requires that two production units produce every part while satisfying Constraints 2 and 3.
In both cases, the initial population of the EA consists exclusively of invalid solutions, i.e., the solutions with \sr\ values less than 1. This is not surprising as the solutions are randomly generated. The optimization process concludes with solutions superior to the randomly generated ones in both the \sr\ and \distI. One can also see that the solutions for the double sourcing scenario have much bigger \distI\ values than the single sourcing solutions. This is also consistent with our expectations, since the double sourcing scenario inherently requires higher transportation demands.
Details of one of the solutions obtained for the double sourcing are presented in Figure~\ref{fig:airbus_decomp-3}, Listing~\ref{list:parts}, Table~\ref{fig:exp1_sites}, and Table~\ref{fig:exp1_countries}.

\begin{table*}[h]
\caption{Results of five independent runs of the evolutionary algorithm for the single sourcing scenario} 
\label{tab:exp1_single}
\centering
\begin{tabular}{ l c c c }
\hline
run & mean initial \sr\ & best initial \distI\ [km] & best final \distI\ [km]\\
\hline
1 & 0.75 & 361531 & 314825 \\
2 & 0.82 & 358380 & \textbf{203101} \\
3 & 0.86 & 297410 & 248772 \\
4 & 0.84 & 264009 & 210220 \\
5 & 0.79 & 273381 & 214430 \\
\hline
\end{tabular}
\end{table*}

\begin{table*}[h]
\caption{Results of five independent runs of the evolutionary algorithm for the double sourcing scenario} 
\label{tab:exp1_double}
\centering
\begin{tabular}{ l c c c }
\hline
run & mean initial \sr\ & best initial \distI\ [km] & best final \distI\ [km]\\
\hline
1 & 0.92 & 799073 & 456173 \\
2 & 0.87 & 706505 & 460592 \\
3 & 0.89 & 705729 & \textbf{425626} \\
4 & 0.86 & 660926 & 469043 \\
5 & 0.91 & 632954 & 471073 \\
\hline
\end{tabular}
\end{table*}

\begin{figure}
\captionsetup{type=listing}
\begin{lstlisting}[escapeinside=`', caption={Assignment of production units to parts for the double sourcing scenario. For each part, two production units are assigned, each of them with a respective production share}, label={list:parts}]
`\includegraphics[height=7.5cm]{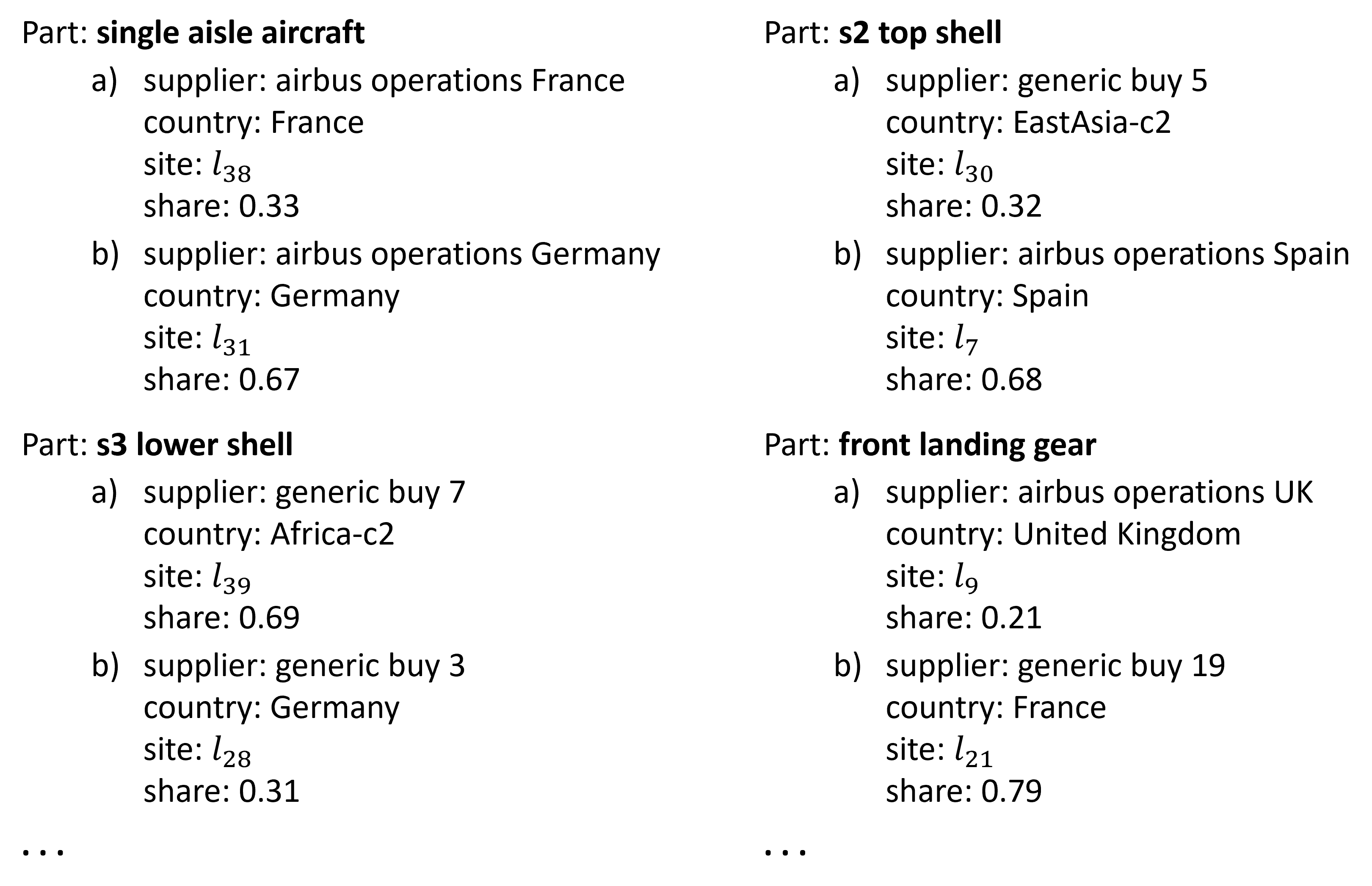}'
\end{lstlisting}
\end{figure}

\begin{table*}[h!]
\centerline{
    \includegraphics[height=9cm]{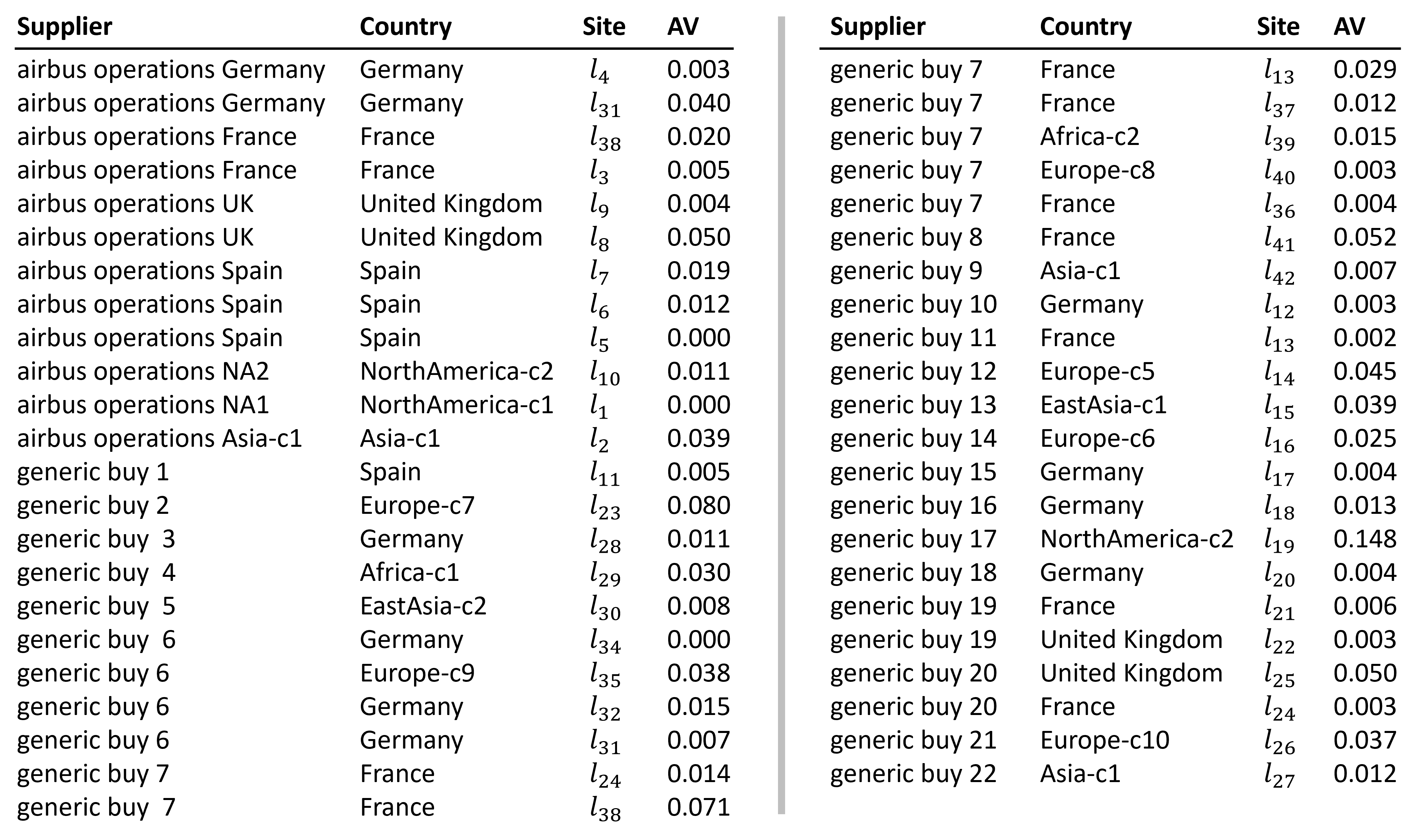}
}
\caption{Distribution of the added value (AV) among suppliers and their production sites.}
\label{fig:exp1_sites}
\end{table*}

\begin{table*}[h!]
    \centering
    \includegraphics[height=7.5cm]{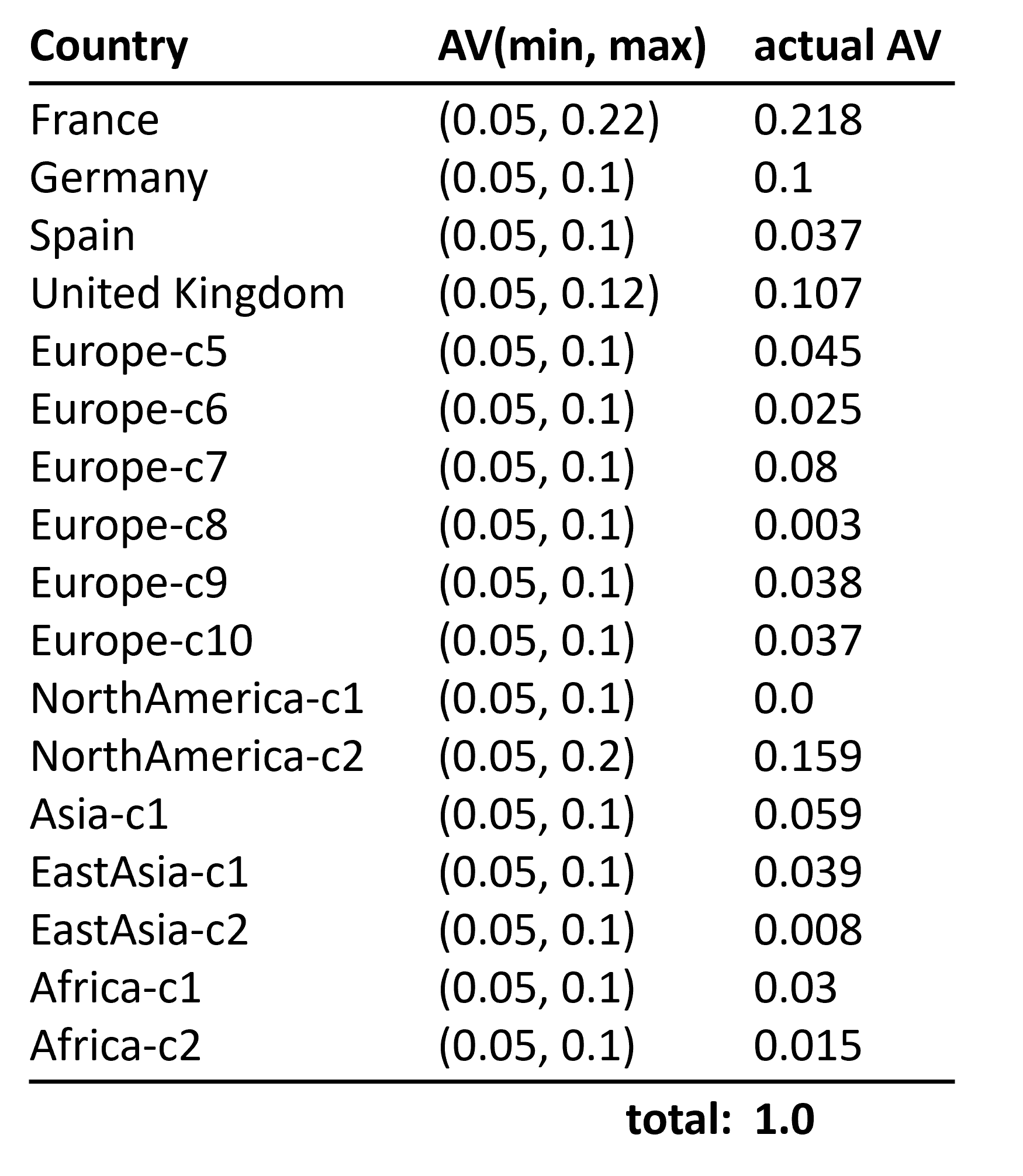}
\caption{Distribution of the added value among countries.}
\label{fig:exp1_countries}
\end{table*}

\subsection{Phase II - Experiments}
\label{sec:experiment2}

The \phaseII\ serves for the final optimization of the industrial system regarding the flows between production sites, aggregating parts in warehouses, and the selection of best-fitting transportation means based on the batching and according to the specified criteria for the optimization.


The bottom part of Fig~\ref{fig:airbus_decomp-3} presents a portion of the transportation graph that was generated based on the outcome from \phaseI. It shows possible connections between production sites and warehouses along with information about transportation means. Information about properties such as the number of parts (i.e., the batch size) to transport, the number of trips with given batches, distance, and duration are not depicted due to the clarity of the figure.

This graph is the input for the optimization using DRAGO algorithm. The results of {\phaseII} are shown in Table~\ref{tab:exp1c}. It summarizes results calculated using one particular output from {\phaseI} and presents KPIs for $CO_2$, duration, distance, and transportation costs optimization. Furthermore, the third and the fourth rows provide duration/$CO_2$ trade-off optimization. The number of produced aircraft was 40 to employ and test batching in the optimization. The comparison of $CO_2$, duration, distance, and transportation costs optimizations with respect to overall transportation distance (resp. duration, $CO_2$, and transportation costs) divided according to transportation means is provided in Table~\ref{tab:exp_comp_dist} (resp. in Table~\ref{tab:exp_comp_dur}, Table~\ref{tab:exp_comp_CO2}, and Table~\ref{tab:exp_comp_costs}).

\begin{table*}[ht]
\caption{{\phaseII} results: Comparison of $CO_2$, duration, distance, and transportation costs optimization with respect to all KPIs (duration, distance, $CO_2$, transportation costs). The number of final products was set to 40 aircraft.} 
\label{tab:exp1c}
\centering
\begin{tabular}{ l c c c c }
\hline
 & \begin{tabular}{@{}c@{}}Duration\\\lbrack{h}\rbrack\end{tabular} & \begin{tabular}{@{}c@{}}Distance\\\lbrack{km}\rbrack\end{tabular} & \begin{tabular}{@{}c@{}}$CO_2$\\emissions \lbrack{g}\rbrack\end{tabular}  & \begin{tabular}{@{}c@{}}Transportation\\costs \lbrack{EUR/km}\rbrack\end{tabular} \\
\hline
$CO_2$ optimization & 62,490 & 1,773,169 & 3.308E+09 & 7,909,327 \\
Duration optimization & 35,199 & 1,309,184 & 11.394E+09 & 13,551,441 \\
Duration/$CO_2$ trade-off:1 & 40,315 & 1,384,028 & 3.483E+09 & 7,846,041 \\
Duration/$CO_2$ trade-off:2 & 36,788 & 1,317,799 & 7.348E+09 & 9,884,600\\
Distance optimization & 37,202 & 1,280,896 & 10,572E+09 & 14,382,365 \\
\begin{tabular}{@{}c@{}}Transportation\\costs optimization\end{tabular} & 52,076 & 1,584,059 & 3.373E+09 & 7,287,918 \\
\hline
\end{tabular}
\end{table*}

\begin{table*}[h!]
\caption{{\phaseII} results: Transportation distances (km) achieved using different optimization strategies (all of which are minimization-based) – $CO_2$ emissions, duration, distance, and transportation costs. Each column represents the distribution of the total distance obtained with the given strategy among the six types of transportation means: large container ship, specialized transport, truck, oversized RoRo ship, oversized Beluga XL, and Kugelbake. The number of final products was set to 40 aircraft.} 
\label{tab:exp_comp_dist}
\centering
\begin{tabular}{ c| c c c c }
\hline
& \multicolumn{4}{c}{Distance [km]}\\
  & $CO_2$ opt. & \begin{tabular}{@{}c@{}}Duration\\opt.\end{tabular} & \begin{tabular}{@{}c@{}}Distance\\opt.\end{tabular}  & \begin{tabular}{@{}c@{}}Transp. costs\\opt.\end{tabular} \\
\hline 
Large Container Ship & 1,020,654.68 & 76,062.67 & 272,914.46 & 831,116.11 \\
Specialized Transp. & 3,145.14 & 3,587.38 & 11,472.92 & 3,145.14 \\
Truck & 304,904,85 & 419,861.12 & 335,622.15 & 438,559 \\
Oversized RoRo Ship & 385,860.09 & 588,882.36 & 457,986.21 & 252,634.47 \\
Oversized Beluga XL & 58,605.07 & 220,790.60 & 202,599.99 & 58,605.07 \\
Kugelbake & 0 & 0 & 300.9 & 0 \\
\hline
\end{tabular}
\end{table*}

\begin{table*}[h!]
\caption{{\phaseII} results: Transportation duration (h) for different optimization strategies (all of which are minimization-based) – $CO_2$ emissions, duration, distance, and transportation costs. Each column represents the distribution of the total transportation duration obtained with the given strategy among the six types of transportation means: large container ship, specialized transport, truck, oversized RoRo ship, oversized Beluga XL, and Kugelbake. The number of final products was set to 40 aircraft.} 
\label{tab:exp_comp_dur}
\centering
\begin{tabular}{ c| c c c c }
\hline
& \multicolumn{4}{c}{Duration [h]}\\
  & $CO_2$ opt. & \begin{tabular}{@{}c@{}}Duration\\opt.\end{tabular} & \begin{tabular}{@{}c@{}}Distance\\opt.\end{tabular}  & \begin{tabular}{@{}c@{}}Transp. costs\\opt.\end{tabular} \\
\hline 
Large Container Ship & 40,826.18 & 3,042.50 & 10,916.57 & 33,244.64 \\
Specialized Transp. & 314.51 & 358.73 & 1,147.29 & 314.51 \\
Truck & 5,822 & 7,889.47 & 6,470.92 & 8,317.88 \\
Oversized RoRo Ship & 15,434.40 & 23,555.29 & 18,319.44 & 10,105.37 \\
Oversized Beluga XL & 93.77 & 353.26 & 324.15 & 93.77 \\
Kugelbake & 0 & 0 & 24.07 & 0 \\
\hline
\end{tabular}
\end{table*}

\begin{table*}[h!]
\caption{{\phaseII} results: $CO_2$ emissions (g) for different optimization strategies (all of which are minimization-based) – $CO_2$ emissions, duration, distance, and transportation costs. Each column represents the distribution of the total $CO_2$ emissions obtained with the given strategy among the six types of transportation means: large container ship, specialized transport, truck, oversized RoRo ship, oversized Beluga XL, and Kugelbake. The number of final products was set to 40 aircraft.} 
\label{tab:exp_comp_CO2}
\centering
\begin{tabular}{ c| c c c c }
\hline
& \multicolumn{4}{c}{$CO_2$ emissions [g]}\\
  & $CO_2$ opt. & \begin{tabular}{@{}c@{}}Duration\\opt.\end{tabular} & \begin{tabular}{@{}c@{}}Distance\\opt.\end{tabular}  & \begin{tabular}{@{}c@{}}Transp. costs\\opt.\end{tabular} \\
\hline 
Large Container Ship & 6.369E+07 & 4,746E+06 & 1.703E+07 & 5.186E+07 \\
Specialized Transp. & 5.661E+07 & 6.457E+07 & 2.065E+08 & 5.661E+07 \\
Truck & 22.733E+07 & 30.833E+07 & 2.526E+08 & 32,499E+07 \\
Oversized RoRo Ship & 6.174E+07 & 9.422E+07 & 7.330E+07 & 4.042E+07 \\
Oversized Beluga XL & 2.899E+09 & 1.092E+10 & 1.002E+10 & 2.899E+09 \\
Kugelbake & 0 & 0 & 248,724 & 0 \\
\hline
\end{tabular}
\end{table*}

\begin{table*}[h!]
\caption{{\phaseII} results: Transportation costs (EUR/km) for different optimization strategies (all of which are minimization-based) – $CO_2$ emissions, duration, distance, and transportation costs. Each column represents the distribution of the total transportation costs obtained with the given strategy among the six types of transportation means: large container ship, specialized transport, truck, oversized RoRo ship, oversized Beluga XL, and Kugelbake. The number of final products was set to 40 aircraft.} 
\label{tab:exp_comp_costs}
\centering
\begin{tabular}{ c| c c c c }
\hline
& \multicolumn{4}{c}{Transportation costs [EUR/km]}\\
  & $CO_2$ opt. & \begin{tabular}{@{}c@{}}Duration\\opt.\end{tabular} & \begin{tabular}{@{}c@{}}Distance\\opt.\end{tabular}  & \begin{tabular}{@{}c@{}}Transp. costs\\opt.\end{tabular} \\
\hline 
Large Container Ship & 1,469,742 & 109,530 & 392,996 & 1,196,807 \\
Specialized Transp. & 874,348 & 997,291 & 3,189,471 & 874,348 \\
Truck & 1,220,366 & 1,610,470 & 1,377,567 & 1,724,537 \\
Oversized RoRo Ship & 2,469,504 & 3,768,847 & 2,932,089 & 1,616,860 \\
Oversized Beluga XL & 1,875,362 & 7,065,299 & 6,483,199 & 1,875,362 \\
Kugelbake & 0 & 0 & 3,944 & 0 \\
\hline
\end{tabular}
\end{table*}

The outcomes of this phase were verified using previously developed simulation model~\citep{Schirrmann2022}.

\section{Conclusions and Further Steps}
\label{sec:conclusions}

This paper introduces a novel approach to optimizing global industrial systems focusing on the aerospace manufacturing sector. Through a two-phase method, it addresses the challenges of designing an efficient industrial system architecture and optimizing the associated transportation network. The integration of evolutionary algorithms and formal optimization techniques demonstrates the ability to handle complex constraints and generate practical solutions.

The proposed framework effectively combines a data-driven model with optimization strategies and it enables the exploration of possible production and logistics configurations. The backbone is represented by OWL ontology, which ensures consistency checks mainly in matching between a part to be produced and available transportation resources. The first phase ensures the creation of a valid production assignment that complies with predefined constraints. The second phase focuses on refining transportation networks and batching strategies to minimize the overall transportation time and the total amount of CO$_2$ generated. 

The experimental results validate the approach's feasibility and are demonstrated by several tested scenarios. The presented approach has been shown to provide a solid foundation for developing resilient and sustainable global industrial systems with potential applications not limited only to aerospace manufacturing. 

Future work will enhance the robustness of the approach, and several future steps are proposed:
\begin{itemize}
    \item Explore simulation-based techniques to complement optimization, particularly for tasks like mixed batching and warehouse utilization.
    \item Integrate advanced optimization techniques, such as multi-objective evolutionary algorithms, to simultaneously address multiple conflicting criteria, including cost, CO2 emissions, and transportation time in the \phaseI.
    \item 
    The algorithm implemented and used in \phaseII\ is sufficient for the problems of this scale. However, an exploitation of algorithms for example from a Linear Optimization (LP)~\citep{bertsimas1997introduction} such as the Network Simplex Method~\citep{cunningham1976network} is planned for following experiments. 
\end{itemize}

\bibliographystyle{elsarticle-harv} 
\bibliography{AirbusDism}

\end{document}